\documentclass[usenames,dvipsnames]{article} 
\usepackage{nips13submit_e,times}
\usepackage{url}

\usepackage{amsmath}
\usepackage{amssymb}
\usepackage{amsfonts}
\usepackage{xcolor}
\usepackage{graphicx}
\usepackage[square, numbers,sort]{natbib}
\usepackage{caption}
\usepackage{subfigure}

\newlength{\imgwidth}
\DeclareMathOperator*{\argmin}{arg\,min}

\renewcommand{\vec}[1]{{\bf #1}}
\newcommand{\Hv}{\vec{H}}
\newcommand{\Vv}{\vec{V}}
\newcommand{\hv}{\vec{h}}
\newcommand{\vv}{\vec{v}}

\newcommand{\data}{\mathbf{D}}
\newcommand{\dv}{\vec{d}}
\newcommand{\dsz}{|\mathbf{D}|}

\newcommand{\pvd}[1][(\Vv=\vv)]{{p_{\data}#1}}
\newcommand{\HHigd}{S(\Hi|\data)}
\newcommand{\HHid}{S_{\data}(\Hi)}
\newcommand{\mihid}{I(\Hi;\data)}
\newcommand{\qdi}[1][\Hv;\data]{\mathrm{AMI}(#1)}

\newcommand{\wij}[1][ij]{w_{#1}}

\newcommand{\partf}{Z}

\newcommand{\ai}[1][i]{a_{#1}}
\newcommand{\bj}[1][j]{b_{#1}}

\newcommand{\phav}{p(\Hv=\hv, \Vv=\vv)}

\newcommand{\phid}[1][\hi]{p_{\data}(\Hi = #1)}

\newcommand{\phigd}[1][\hi]{p(\Hi = #1|\Vv=\dv)}

\newcommand{\tr}[1]{{#1}^\mathrm{T}}

\newcommand{\hi}[1][i]{h_{#1}}
\newcommand{\vj}[1][j]{v_{#1}}
\newcommand{\sj}[1][j]{\sigma_{#1}}
\newcommand{\nvj}[1][j]{\frac{\vj[#1]}{\sj[#1]}}

\newcommand{\Hi}[1][i]{H_{#1}}
\newcommand{\Vj}[1][j]{V_{#1}}

\newcommand{\refeq}[1]{Eq.~\ref{#1}}
\newcommand{\reffig}[1]{Fig.~\ref{#1}}
\newcommand{\refFig}[1]{Figure~\ref{#1}}
\newcommand{\reftab}[1]{Table~\ref{#1}}

\newcommand{\ts}{\rho}

\newcommand{\tee}[1]{\mathrm{AMI}\left(#1\right)}
\newcommand{\rli}[1]{\overline{\mathrm{AMI}}\left(#1\right)}
\newcommand{\rlii}[1]{\widetilde{\mathrm{AMI}}\left(#1\right)}

\newcommand{\sign}{\mathrm{Sign}}

\title{Approximated Infomax Early Stopping:\\
Revisiting Gaussian RBMs on Natural Images}

\author{
Taichi Kiwaki\\
The University of Tokyo\\
\texttt{kiwaki@sat.t.u-tokyo.ac.jp}
\And
Takaki Makino\\
The University of Tokyo\\
\texttt{mak@sat.t.u-tokyo.ac.jp}
\And
Kazuyuki Aihara\\
The University of Tokyo\\
\texttt{aihara@sat.t.u-tokyo.ac.jp}
}
\nipsfinalcopy

\begin{document}

\maketitle

\begin{abstract}
 We pursue an early stopping technique that helps Gaussian Restricted Boltzmann Machines (GRBMs) to gain good natural image representations in terms of overcompleteness and data fitting. GRBMs are widely considered as an unsuitable model for natural images because they gain non-overcomplete representations which include uniform filters that do not represent useful image features. We have recently found that GRBMs once gain and subsequently lose useful filters during their training, contrary to this common perspective. We attribute this phenomenon to a tradeoff between overcompleteness of GRBM representations and data fitting. To gain GRBM representations that are overcomplete and fit data well, we propose a measure for GRBM representation quality, approximated mutual information, and an early stopping technique based on this measure. The proposed method boosts performance of classifiers trained on GRBM representations. 
\end{abstract}


\section{Introduction}	

While the Restricted Boltzmann Machines (RBMs) have been demonstrated to be effective in various tasks \cite{Larochelle:2008wl, Nair:2008tv, Anonymous:vtpM00Dt, Scholkopf:2006vl}, they often fail to learn overcomplete representations from continuous data, especially from natural images \cite{Courville:2011vh, Ranzato:2010vs}. For example, RBMs trained on natural images learn a large number of uniform filters that do not represent sharp edges. The failure of RBMs in the application for continuous data has been attributed to the deficiency of Gaussian RBMs (GRBMs), a variant of RBMs for continuous data, in capturing the covariances between data variables \cite{Ranzato:2010vs}. This perspective has led people to several model extensions of RBMs that can model data covariances \cite{Courville:2011vh, Dahl:2010uy, Ranzato:2010vs}. 

\label{qr:sec}
\begin{figure}[h]
\centering
\begin{picture}(300,200)
\put(-40,120){\includegraphics[width=0.18\imgwidth,bb=10 -30 210 270]{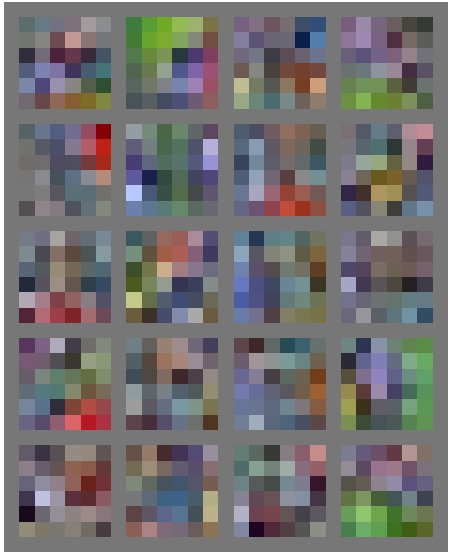}}
\put(35,120){\includegraphics[width=0.18\imgwidth,bb=10 -30 210 270]{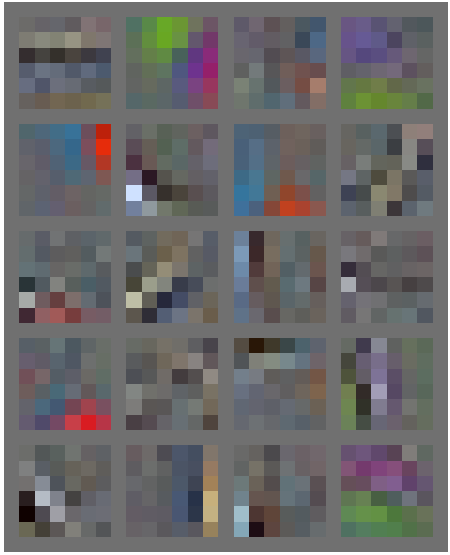}}
\put(-40,20){\includegraphics[width=0.18\imgwidth,bb=10 -30 210 270]{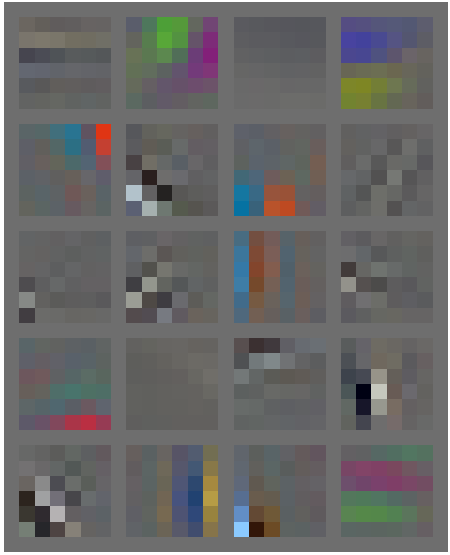}}
\put(35,20){\includegraphics[width=0.18\imgwidth,bb=10 -30 210 270]{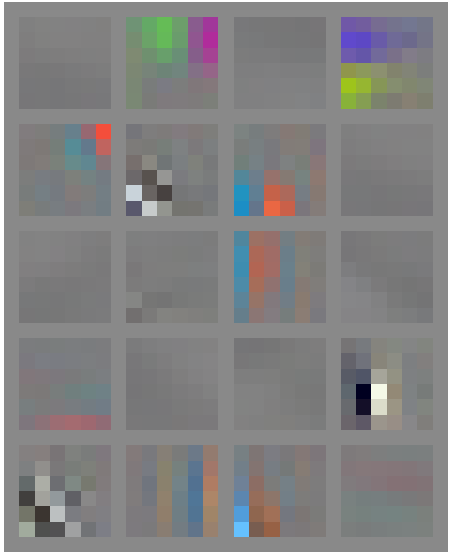}}
\put(-18,115){{\large \bf (a)}}
\put(57,115){{\large \bf (b)}}
\put(-18,15){{\large \bf (c)}}
\put(57,15){{\large \bf (d)}}
\put(168,-50){\includegraphics[angle=90,height=0.65\imgwidth,bb=50 150 550 500]{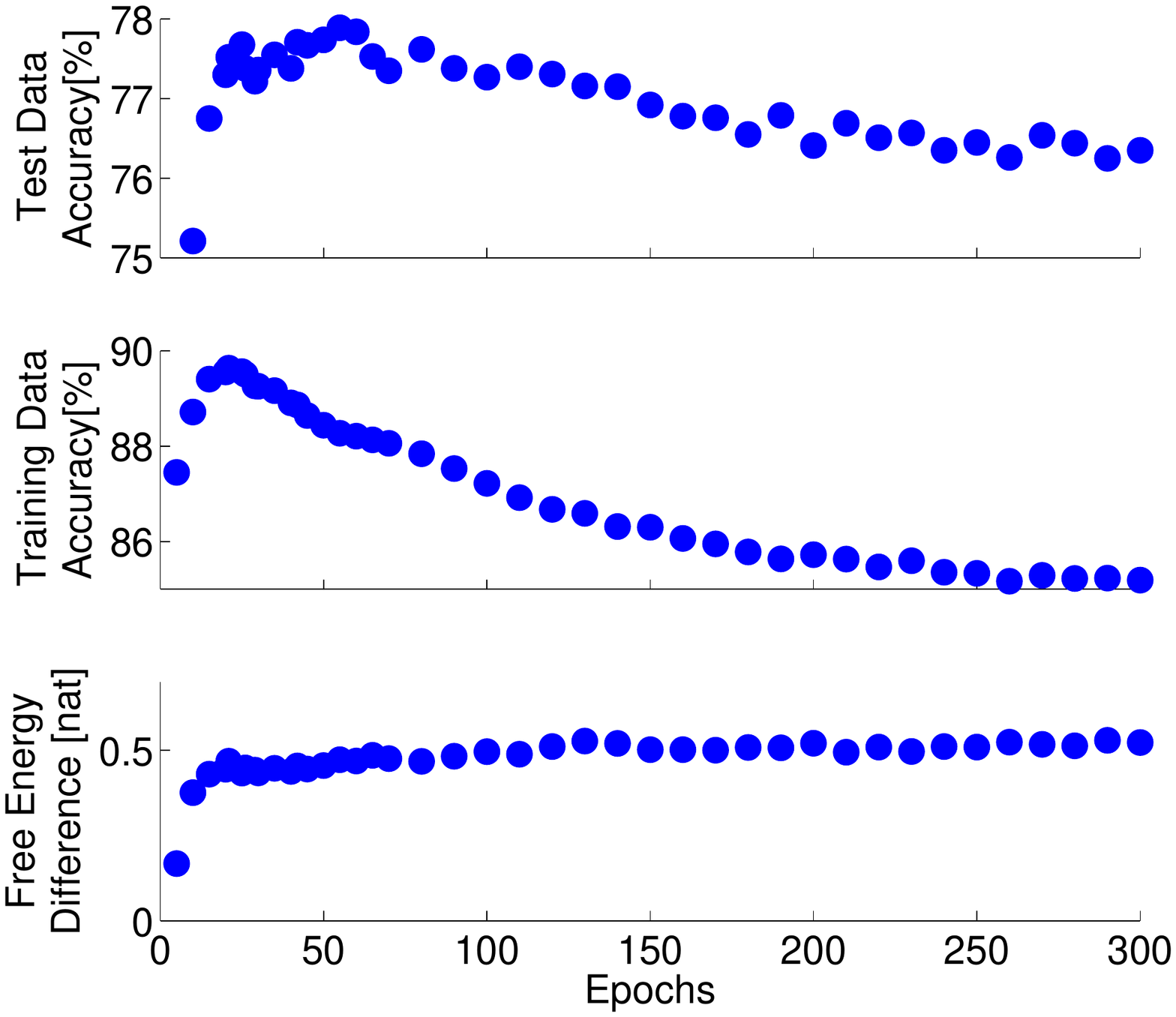}}
\put(115,160){{\large \bf (e)}}
\put(115,105){{\large \bf (f)}}
\put(115,50){{\large \bf (g)}}
\end{picture}
\vspace{-5mm}
 \caption{from {\bf (a)} to {\bf (d)}: Development of feature representations. Shown are 20 randomly selected filters of a GRBM after {\bf (a)} 10 epochs, {\bf (b)} 30 epochs, {\bf (c)} 100 epochs, and {\bf (d)} 300 epochs of training. {\bf (e)} and {\bf (f)}: Duration of RBM training and classification performance. {\bf (g)} Development of FED. We used GRBM training setting (d) and SVM parameters for fig. 4.}
 \label{accepoch:fig}
\end{figure}


It is, however, not widely known that GRBMs once gain useful filters and subsequently lose many of them during the training \cite{Kiwaki:2013}. Figure~\ref{accepoch:fig}~(a) to (d) show filters learned by a GRBM from CIFAR-10 in different stages of the training. As is reported \cite{Courville:2011vh, Ranzato:2010vs}, the GRBM gains a lot of uniform filters at the end of the training (after 300 sweeps over training data, or epochs) in \reffig{accepoch:fig}~(d). However, \reffig{accepoch:fig}~(b) and (c) suggest us that these uniform filters are actually degenerated from meaningful filters that represented edges or color gradients in the middle stages of the training (after 30 and 100 epochs). As apparent, we cannot trace back the GRBM training process for a useful representation into the early stages of the training where all the filters were nearly noisy initial states as in \reffig{accepoch:fig}~(a). 

The change in the quality of GRBM feature representations impacts on the performance of a classifier trained on the representations. Here, we consider a stack of the GRBM and an L2-SVM with the linear kernel, and see classification performance on CIFAR-10. Figure~\ref{accepoch:fig}~(e) and (f) illustrate the relationship between duration of the GRBM training and classification performance of the stacked system. The percentage accuracy on both test and training data increases in the early stages of the training (1--30 epochs) and then declines in the late stages of the training (30--300 epochs), corresponding to the quality of GRBM feature representations. 
 
These phenomena cannot be simply explained that the GRBM overfits the training data, owing to two reasons. First, decline in the percentage accuracy is observed on both the test and training data. However, GRBM overfitting would result in decline only in the test data accuracy and increase in the training data accuracy. Second, the difference between the averaged free energy of test and training data (hereafter, free energy difference, or FED), a common overfitting measure for RBMs \cite{Hinton:2012un}, remains small throughout the GRBM training as in \reffig{accepoch:fig}~(g). If GRBM overfitting were responsible for the phenomena, FED would show a significant increase as the GRBM loses its useful filters. 


Compared with our previous study \cite{Kiwaki:2013}, there are two contributions to report. First, to better understand the phenomena, we claim a detailed mechanism on how GRBMs once gain and eventually lose useful filters during the training (Section~3). We there argue that GRBMs lose overcomplete representations because the maximum likelihood estimation of GRBMs can penalize such representations. This argument leads us an idea of a tradeoff between overcompleteness of GRBM representations and fitting of GRBMs to training data, which suggests us that better fitting of a GRBM does not always result in better representations even if the GRBM does not overfit. The tradeoff between overcompleteness and fitting motivates us to perform early stopping training for GRBMs. 

Second, we propose an efficient early stopping technique for GRBM representations that are overcomplete and fit data well. 
 In Section~4, We describe our proposed method, dubbed a-infomax early stopping for GRBMs, that maximizes approximated mutual information (AMI), which is the sum of mutual information between each hidden unit and data. A-infomax early stopping is efficient while alternative strategies, such as directly using test or validation data accuracy at each candidate of stopping time, can incur us huge computational costs in performing supervised training and convolutional feature extraction. Moreover, parameter tuning for a-infomax early stopping is easier than our previous approaches \cite{Kiwaki:2013}. In Section~5, we present experiments to explore the benefits from a-infomax early stopping. 

\section{Gaussian RBMs}
RBMs are a Markov random field of a bipartite graph that consists of two layers of variables: visible layer representing data, and hidden layer representing latent variables. GRBMs are one of the variants of RBMs with real-valued data and binary latent variables, whose joint probabilities are 
\begin{align}
 &\phav = \frac{1}{\partf}\exp\left(
\sum_{i=1}^{M}\sum_{j=1}^N \nvj \wij \hi + \sum_{i=1}^M \ai\hi -\sum_{j=1}^N\frac{(\vj-\bj)^2}{\sj^2}
\right),\label{rbm:eq}
\end{align}
where $M$ and $N$ are the number of hidden and visible variables, $\Hv$ and $\Vv$ are a vector notation of hidden and visible variables, i.e., $\Hv = \tr{(\Hi[1], \dots, \Hi[M])}$ and $\Vv = \tr{(\Vj[1], \dots, \Vj[N])}$, $Z$ is the partition function, $\wij$ are connection weights between $\Hi$ and $\Vj$, $\ai$ and $\bj$ are biases for hidden and visible variables, $\sj$ are parameters that control the variances of visible variables, and $\hv$ and $\vv$ are arbitrary realizations of $\Hv$ and $\Vv$ with $\hi$ and $\vj$ being their elements, i.e., $\hv = \{0,1\}^M$ and $\vv \in \mathbb{R}^{N}$. 
 A GRBM models a data distribution as a Gaussian mixture distribution, in which the number of the mixing components is exponential in the number of the hidden variables.


\subsection{Training RBMs}
 GRBMs are trained by adjusting their parameters $\wij$, $\ai$, and $\bj$ so that the log-likelihood of data is maximized. The weight parameter update rule for GRBMs can be derived by taking the gradient of the log-likelihood:
\begin{align}
  \frac{\partial \log p}{\partial \wij} &\propto \mathbb{E}_{\pvd[]}[\Hi\Vj] - \mathbb{E}[\Hi \Vj],\label{updt:eq}
\end{align}
where $\mathbb{E}_{\pvd[]}[\cdot]$ denotes an expectation by an empirical distribution $\pvd$. The first term of the r.h.s. of \refeq{updt:eq} is called the positive gradient and the second term is called the negative gradient. Because exact inference of the negative gradient is intractable, practical applications use approximated inference based on log-likelihood approximation, such as contrastive divergence \cite{Hinton:2002wb} or persistent contrastive divergence \cite{Tieleman:2008gw}. 
 Weights are often updated by a gradient that is averaged over a set of several training cases, which is called a batch. We define an epoch as a unit duration of RBM training in which all the training cases in a dataset are used once for weight updates.

 Sparsity helps unsupervised learning systems learn suitable data representations \cite{Lee:2009tm,Lee:2009wl,Coates:2011wo}. GRBMs can obtain sparse representations by sparse regularization with two control parameters: the regularization strength $\lambda$ and the sparsity target $\ts$. This regularization constrains activation probabilities of GRBM hidden units to be close to $\ts$ \cite{Lee:2009tm}. 

\subsection{GRBMs as Data Encoders}
There are several ways to obtain data representations using GRBMs. The natural choice is the conditional expectation of hidden units which can be efficiently computed, but this is not necessarily the best strategy \cite{Coates:2011uda}. We can gain alternative representations based on various encoding schemes. Particularly when using SVMs in supervised learning, as \citet{Coates:2011uda} report, an encoding scheme called ``soft thresholding'' is effective for achieving high classification accuracy. 

\section{Mechanisms of GRBMs Losing Useful Representations}
\label{evaluating:sec}

\begin{figure}[h]
\centering
\begin{picture}(400,130)
\put(0,30){\includegraphics[width=0.27\imgwidth,bb=0 -20 300 475,angle=0]{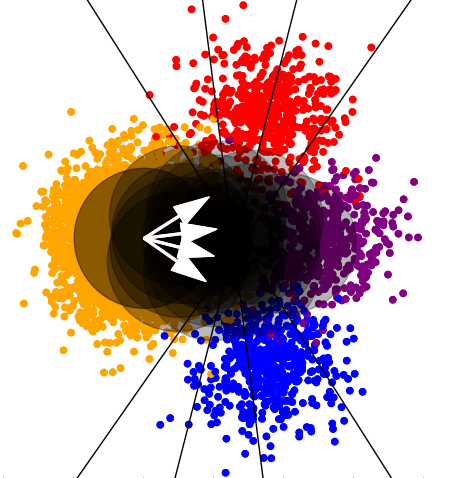}}
\put(85,30){\includegraphics[width=0.27\imgwidth,bb=0 -20 300 475,angle=0]{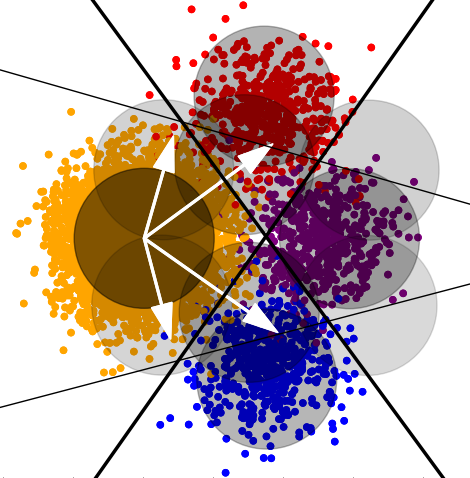}}
\put(180,30){\includegraphics[width=0.27\imgwidth,bb=0 -20 300 475,angle=0]{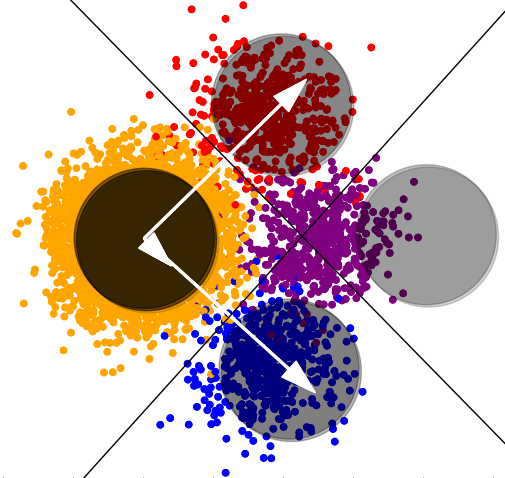}}
\put(0,125){{\large \bf A}}
\put(285,125){{\large \bf B}}
\put(280,20){\includegraphics[width=0.24\imgwidth,bb=-80 0 390 500]{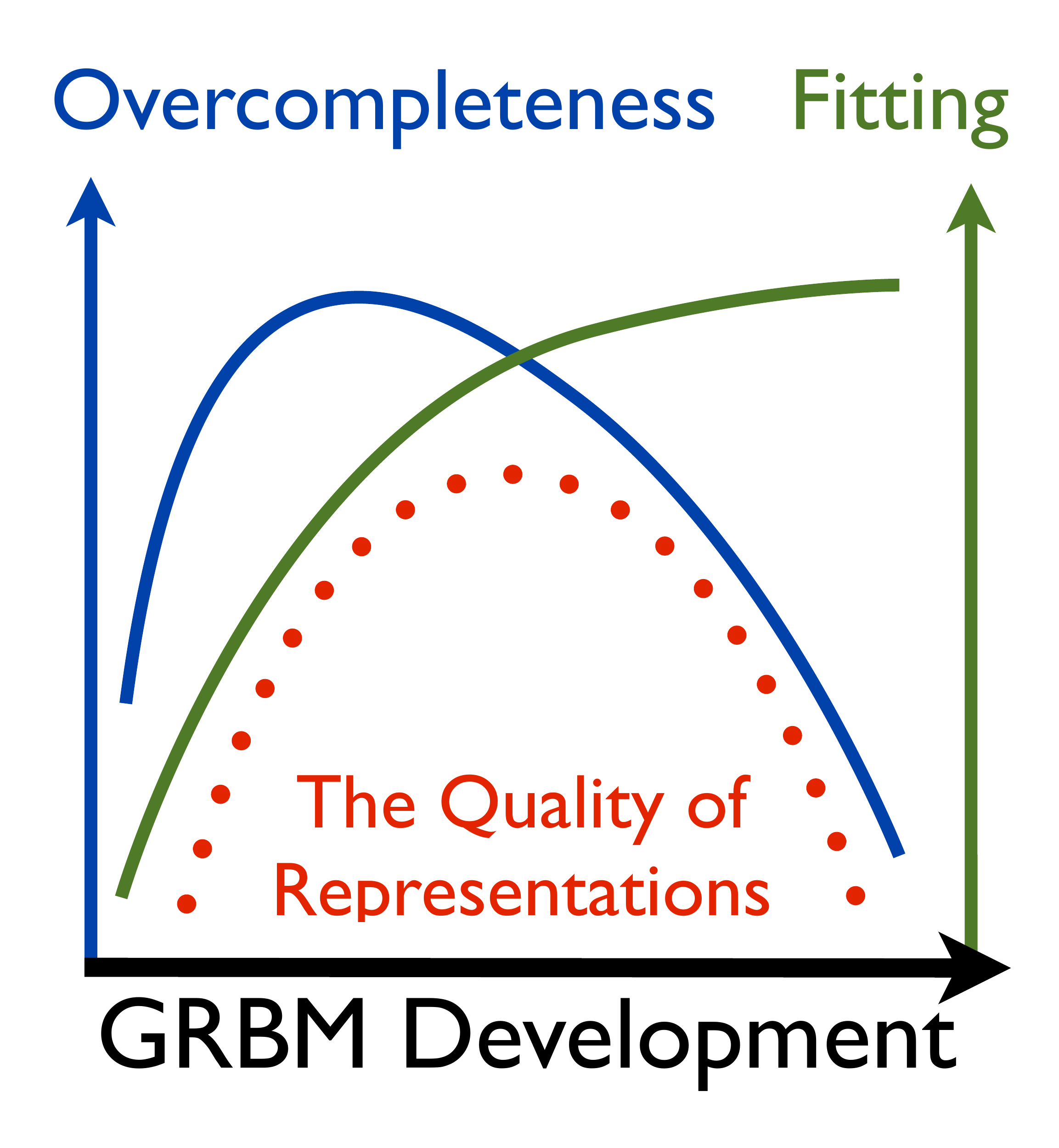}}
\put(45,25){{\large \bf (a)}}
\put(132,25){{\large \bf (b)}}
\put(235,25){{\large \bf (c)}}
\end{picture}
\vspace{-13mm}
 \caption{{\bf A}: GRBM filter number adjustment. State of the GRBM {\bf (a)} after 10 epochs, {\bf (b)} after 140 epochs, {\bf (c)} after 2000 epochs. Points represent the data where colors indicate components of the mixture, white arrows represent GRBM filters, black lines represent planes that the activation probabilities of GRBM hidden units are at 0.5, and gray shadows represent the GRBM components (area within 2$\sigma$) where opacity corresponds to their proportion in the mixture. The GRBM was trained using the true likelihood gradient. {\bf B} Tradeoff between overcompleteness and fitting.}
 \label{mechanism:fig}
\end{figure}
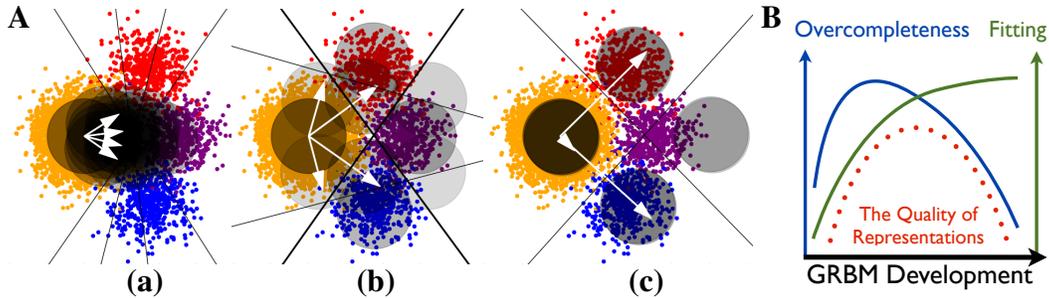
\vspace{-2mm}

As we have seen in Section~1, a GRBM gains a lot of meaningful filters after a certain period of training, but then loses a large number of them if the training continues. Such representations gained after prolonged training are widely regarded as a poor representation because they are not overcomplete \cite{Courville:2011vh, Ranzato:2010vs}. Representations are said overcomplete if there are more filters (of course, it is assumed in literature that the filters are in mutually different directions, and uniform filters are therefore excluded) than the effective dimensionality of data \cite{Olshausen:1997uh}. 

We view that the lost of GRBM overcompleteness arises from adjustment of the number of non-uniform GRBM filters according to the effective data dimensionality. What we assume here is that a data distribution on a low dimensional manifold is close to the family of distributions that GRBMs with a small number of hidden units --- as small as the manifold dimensionality --- can express. In maximum likelihood learning of a GRBM with a large number of hidden units, excess filters are thus attenuated to reduce negative effects that those filter can have on the likelihood. 

To verify that GRBM filter number adjustment is responsible for the lost of overcompleteness, we performed a toy data experiment on a GRBM with four filters. Data was generated from a two dimensional Gaussian mixture with four components: one dense component and three sparse components that are the same distance apart from the dense one. This distribution can be well approximated, but cannot be completely expressed by a GRBM with two filters. 

The GRBM develops as follows. The GRBM initially gains a representation that is overcomplete, but does not fit data well as in \reffig{mechanism:fig}~A~(a). After 130 more epochs of training, the GRBM better fits the data and GRBM components roughly spread over the data points as in \reffig{mechanism:fig}~A~(b). 
 After 2000 epochs of training, the GRBM further fits the data and concisely assigns its components to the four mixture components of the data distribution as in \reffig{mechanism:fig}~A~(c).  However, overcompleteness is damaged because two filters are attenuated to zero.  One important point in this experiment is that the GRBM does not overfit the training data; the averaged log-likelihood of another data generated from the data distribution will remain close from that of the training data. 

The balance between data fitting and overcompleteness dominates the performance of a linear classifier (e.g., an SVM) trained on the GRBM representation for a component classification problem. After 10 epochs of training, the classification performance is expected to be low because planes that separate regions in which a hidden unit is likely 0 or 1 do not clearly divide the data components (in \reffig{mechanism:fig}~A~(a)). 
 After 140 epochs of training, the classification performance will become higher because two separation planes (denoted by thick lines) divide all the data components, and the classifier can assign data labels based on the configuration of two hidden units that correspond to those planes. After 2000 epochs of training, however, the performance will be degraded because two effective separation planes fail to divide one data component (in purple) from the others. 


Description above suggest us that two important aspects of the quality of representations, overcompleteness and fitting, are in a tradeoff relation in GRBM training, except for early stages of training where all the filters are nearly zero. Figure~\ref{mechanism:fig}~B sketches this concept. GRBM overcompleteness (the blue curve) rapidly grows in the early stages of training, but eventually becomes damaged as fitting of a GRBM to training data (the green curve) improves. The quality of GRBM representations is first improved and then damaged according to this balance between fitting and overcompleteness. 
 This perspective also tells us that higher data likelihood of GRBMs does not always result in better representations. We believe that our perspective is related to reports that maximum likelihood training of RBMs for lower layers of deep networks is suboptimal \cite{Luo:2013vx, LeRoux:2008vc}, and therefore foresee that the tradeoff can be observed in maximum likelihood training of other RBM variants. 


\section{Approximated Mutual Infomax Early Stopping for Better GRBM Representations}
\label{ES:sec}
The perspective that we have described so far motivates us to perform early stopping in GRBM training for better representations in terms of fitting and overcompleteness. Because the standard early stopping is a method to prevent overfitting \cite{Amari:1997ej}, we here explore a new early stopping criterion. 

We now present two candidate criteria. First candidate is validation data accuracy, the same criterion for the standard method. It is straightforward to use validation data accuracy for early stopping because it is exactly what we want to optimize. However, validation data accuracy is computationally expensive because of supervised training phase where we compute feature maps by convolving GRBM filters and input images, and then train SVMs on a huge set of feature maps \cite{Coates:2011uda, Courville:2011vh}. 

Second candidate is the information of data held by GRBM representations, which directly measures the GRBM representation quality. We here propose an infomax approach \cite{Bell:1995vn} with a novel information measure, approximated mutual information (AMI) defined as:
\begin{align}
\qdi \triangleq \sum_{i=1}^{M}\mihid. \label{approx:eq}
\end{align}
This quantity approximates the mutual information between data and whole hidden variables $I(\Hv;\data)$. Actually, we can verify that AMI of GRBMs is an upper bound of the mutual information, i.e., $I(\Hv;\data)\le \qdi$ by simply considering $S_{\data}(\Hv)\le\sum_i^N S_{\data}(\Hi)$ where $S$ denotes entropy. The idea is to perform early stopping at the timing where AMI reaches its maximum. Thus, we call our method approximated-infomax early stopping, more shortly, a-infomax early stopping. 

AMI can be efficiently computed for GRBMs. To begin with, the mutual information between each hidden unit and data can be decomposed as $\mihid = \HHid - \HHigd$, where $\HHid$ and $\HHigd$ are entropy and conditional entropy of hidden units when they are used to encode data. Both of these entropies can be efficiently computed. 
First, $\HHid$ can be directly computed from the activation probabilities of hidden units when they encode data, \begin{align}
\phid &= \sum_{\dv\in\data} \phigd \frac{1}{\dsz}. 
\end{align}
 Second, $\HHigd$ can be computed as 
\begin{align}
\HHigd &=  \frac{1}{\dsz}\sum_{\dv\in\data} \sum_{\hi\in\{0,1\}}\phigd\log\frac{1}{\phigd},
\label{dcue:eq}
\end{align}
where conditional probabilities are efficiently computed because RBM hidden units are conditionally independent given a data vector. For further technical details on a-infomax early stopping, see the appendix. Note that AMI is computed within time linear in data size without sampling. AMI does not require supervised training phase with convolutional feature extraction and SVM training. 

There are two intuitive interpretations of a-infomax early stopping. The first interpretation is that the mutual information between data and each hidden variable is a good indicator for non-uniformity of the corresponding filter. Suppose a GRBM filter for a sharp edge. The hidden variable corresponding to this filter shows strong correlation between the input image patches to the GRBM visible layer. The mutual information therefore become greater than zero. On the other hand, the mutual information becomes nearly zero with uniform filters, because the corresponding hidden units are almost statistically independent of the input images. AMI, the sum of the mutual information over the hidden variables, thus captures the sharpness of the filters throughout the whole representation. 

The second interpretation is that AMI roughly measures overcompleteness of representations. Let us consider an extreme example where we add an extra hidden unit $\hat{H}$ to a perfect data representation $\Hv$, such that $\data = f(\Hv)$ with $f(\cdot)$ being a function. Introduction of this new variable helps better overcompleteness of the resulting representation $\{\Hv, \hat{H}\}$ if that variable reflects information about the data, $I(\hat{H};\data)>0$. AMI is sensitive to this improvement and increases as $ \qdi[\{\Hv, \hat{H}\};\data] = \qdi + I(\hat{H};\data)$. On the other hand, mutual information does not reflect the improvement in overcompleteness and remains the same, $I(\{\Hv, \hat{H}\};\data) = I(\Hv;\data)$. Therefore, AMI can be considered as a corrected mutual information for measuring overcompleteness. 


\section{Experiments and Discussion}
\label{exp:sec}

\subsection{AMI and The Quality of GRBM Representations}

\label{AIQR:sec}
\begin{figure}[t]
\begin{picture}(300,210)
\put(20,95){\includegraphics[width=.99\imgwidth,bb=100 12 1100 300]{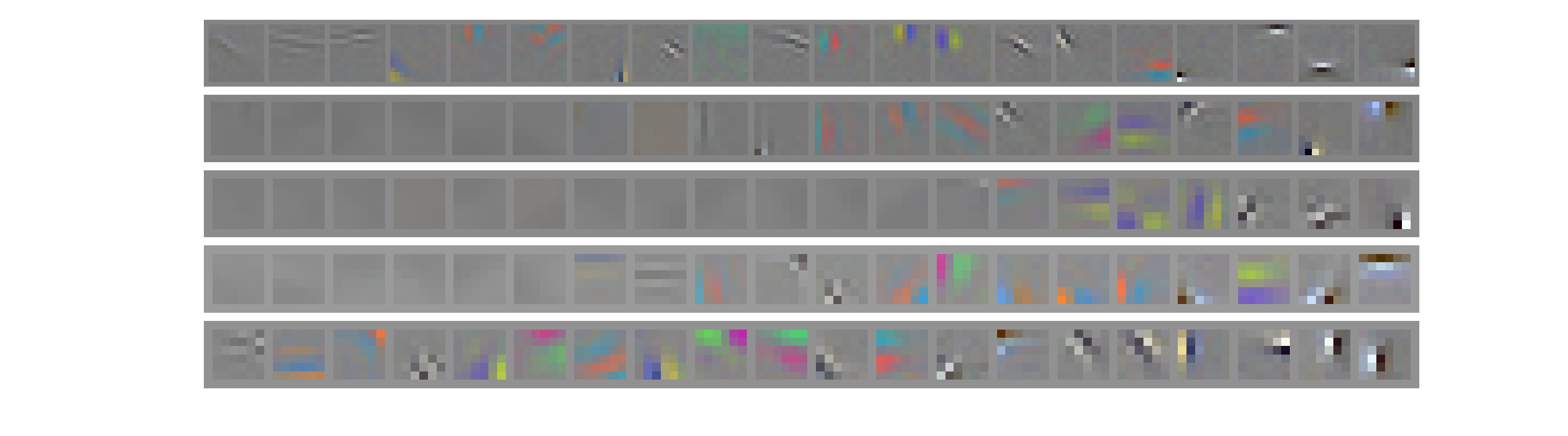}}
\put(20,-5){\includegraphics[width=\imgwidth,bb=48 0 555 300]{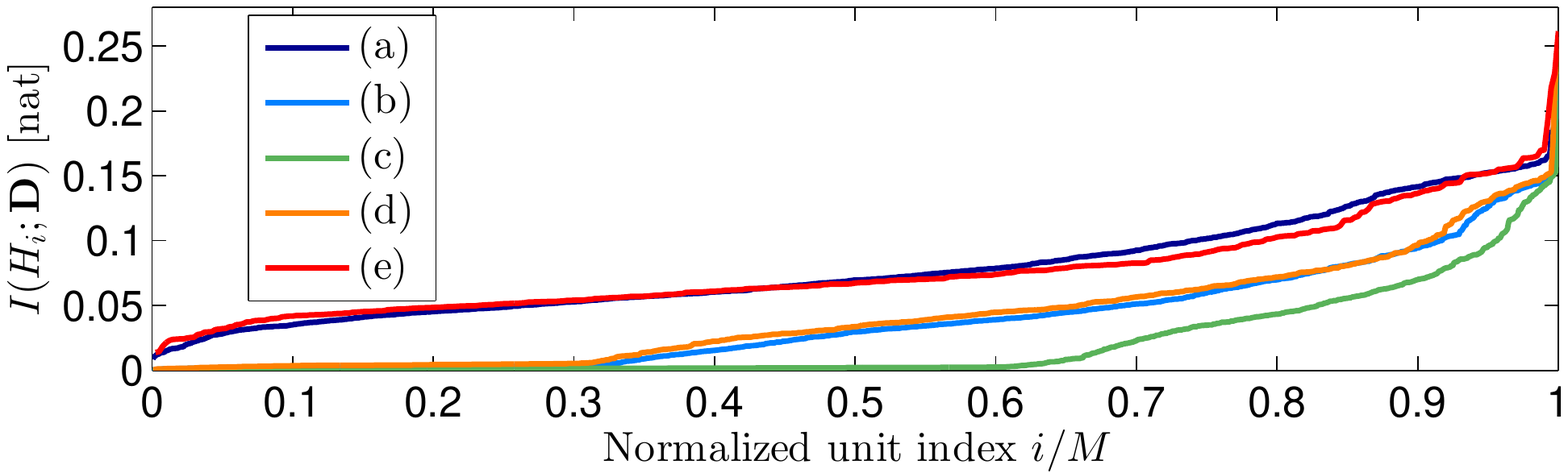}}
\put(0,185){{\Large \bf A}}
\put(20,185){{\large  (a)}}
\put(20,166){{\large  (b)}}
\put(20,147){{\large  (c)}}
\put(20,128){{\large  (d)}}
\put(20,109){{\large  (e)}}
\put(0,90){{\Large \bf B}}
\linethickness{1.5pt}
\multiput(44,22)(5,0){62}{\line(0,1){1.5}}
\multiput(42,20)(0,5){36}{\line(0,1){1.5}}
\multiput(350,20)(0,5){36}{\line(0,1){1.5}}
\color{Green}\multiput(255,15)(0,15){13}{\line(0,1){10}}
\color{orange}\multiput(160,15)(0,15){13}{\line(0,1){10}}
\color{Cerulean}\multiput(180,15)(0,15){13}{\line(0,1){10}}
\color{red}\multiput(50,15)(0,15){13}{\line(0,1){10}}
\color{blue}\multiput(55,15)(0,15){13}{\line(0,1){10}}
\end{picture}    
 \caption{{\bf A}: GRBM representations under various training settings. All filters were gained after 80 epochs of training. The hidden unit index $i$ was put in ascending order of the mutual information. 20 filters with $i/M=0.025, 0.075, 0.125, \dots, 0.975$ are shown for parameter settings (a) \{1600, 432, 0.003, PCD\}, (b) \{1600, 192, 0.003, PCD\}, (c) \{1600, 108, 0.003, PCD\}, (d) \{800, 108, 0.003, PCD\}, and (e) \{400, 108, 0.003, PCD\}.
 {\bf B}: Mutual information between each hidden unit and data. 
 Vertical lines indicate the points where the plots of mutual information in the corresponding color exceeds $0.02[\mathrm{nat}]$. }
\label{imgRFs:fig}
\end{figure}

In this section, we investigate the relationship between AMI and the quality of GRBM representations under various settings of GRBM training: the weight update algorithms and GRBM hyper parameters. For the algorithms, we examined persistent contrastive divergence (PCD) \cite{Tieleman:2008gw} and contrastive divergence (CD) \cite{Hinton:2002wb}. For hyper parameters, there are four major GRBM parameters to be considered: the learning rates, the batch size, and the number of hidden and visible units. However, we omit batch size from our investigation because increasing the batch size has similar effects as lowering the learning rate for sufficiently large batch sizes. Therefore, we here focus on the number of hidden and visible units, and the learning rate. 

 We trained GRBMs by PCD on 100,000 image patches from CIFAR-10 \cite{Krizhevsky:2009tr}, which were preprocessed by contrast normalization and whitening. We first performed 5-fold cross validation to determine the optimal parameters for GRBM sparsity, feature encoders, and L2-SVMs. L2-SVMs were trained on GRBM representations pooled over quadrants of CIFAR images to compute validation data accuracy. 
 After finding the best parameters, we trained and evaluated GRBMs on the standard training and test data of CIFAR-10. GRBMs were trained with different numbers of hidden and visible units (or the size of receptive fields), the learning rate, and the weight update algorithm. To describe the GRBM training setting, we use a tuple \{$M$, $N$, $\delta$, algorithm\} where $\delta$ denotes the learning rate. We here examined seven GRBM training settings from (a) to (g): from (a) to (c), we varied the receptive field size, and from (c) to (e), we varied the number of hidden units. In (f), we tried a lower learning rate. In (g), we used CD instead of PCD. We trained GRBMs for 80 epochs in all the trials except for (f) where the training duration was 240 epochs. 


To begin with, let us discuss how mutual information between data and each hidden variable relates to the sharpness of the corresponding filter, and how AMI relates to the overcompleteness of a whole representation. 
 \refFig{imgRFs:fig} {\bf A} (a) to (e) show uniformly taken samples of GRBM filters which are in ascending order of the mutual information $\mihid$. From the vertical lines in \refFig{imgRFs:fig}, it is clearly seen that hidden units with sharp edge filters hold relatively large values of $\mihid$ (larger than $0.02[\mathrm{nat}]$). 
Finally, let us turn to whole representations from individual filters. \refFig{imgRFs:fig} {\bf B} shows plots of $\mihid$ where each curve corresponds to the training settings from (a) to (e). AMI of a representation is the product of the number of hidden units and the area between the corresponding curve and the horizontal axis in \reffig{imgRFs:fig} {\bf B}. As is obvious, overcomplete representations with a large number of non-uniform filters show a large value of AMI. 

Moreover, two interesting trends can be seen. First, the larger the number of hidden units becomes, the worse overcompleteness becomes. This can be explained as the number of excess filters in terms of likelihood maximization increases as the total number of hidden units becomes large. Second, the larger the number of visible units becomes, the better overcompleteness becomes. This can be attributed to a large effective data dimensionality. Large number of visible dimensions helps data variations to lie in orthogonal directions. This results in a large effective data dimensionality, and therefore inhibits the lost of overcompleteness.

\begin{figure}[t]
\centering
\begin{tabular}{cc}
\begin{minipage}{0.44\textwidth}
\includegraphics[width=0.5\imgwidth,bb=0 200 500 600]{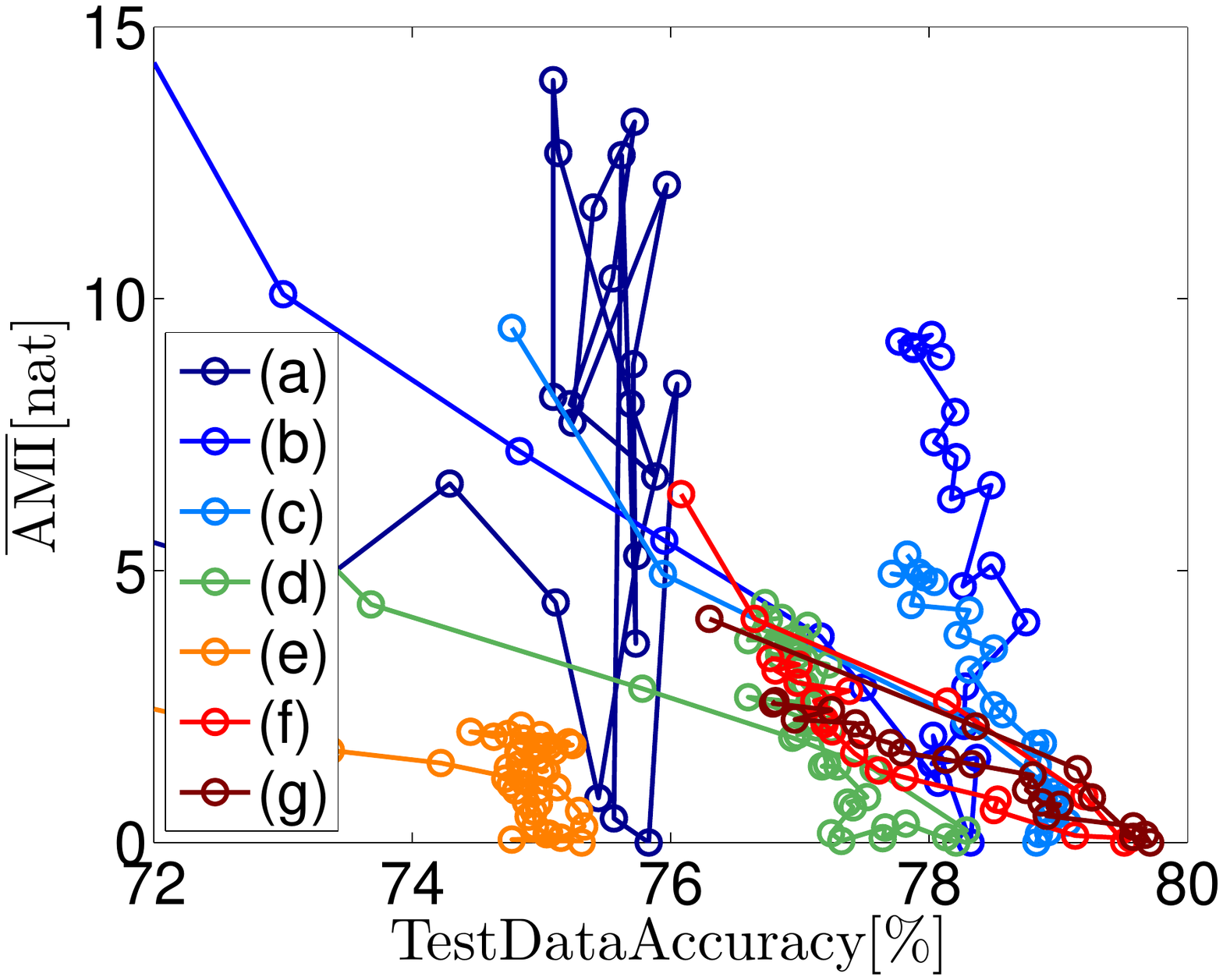}
\vspace{1mm}
 \caption{Performance and $\overline{\mathrm{AMI}}$ (See appendix for the definition). The parameter settings (a) to (e) are the same as in \reffig{imgRFs:fig}. Here, we also show settings (f) \{1600, 108, 0.001, PCD\} and (g) \{1600, 108, 0.003, CD\}
. 
Series of points in a color correspond to each run of GRBM training and lines indicate the temporal order.}
\label{ADI_PFM:fig}
\end{minipage}
\hspace{4mm}
\begin{minipage}{0.44\textwidth}
\includegraphics[width=0.7\imgwidth,bb=20 60 530 300]{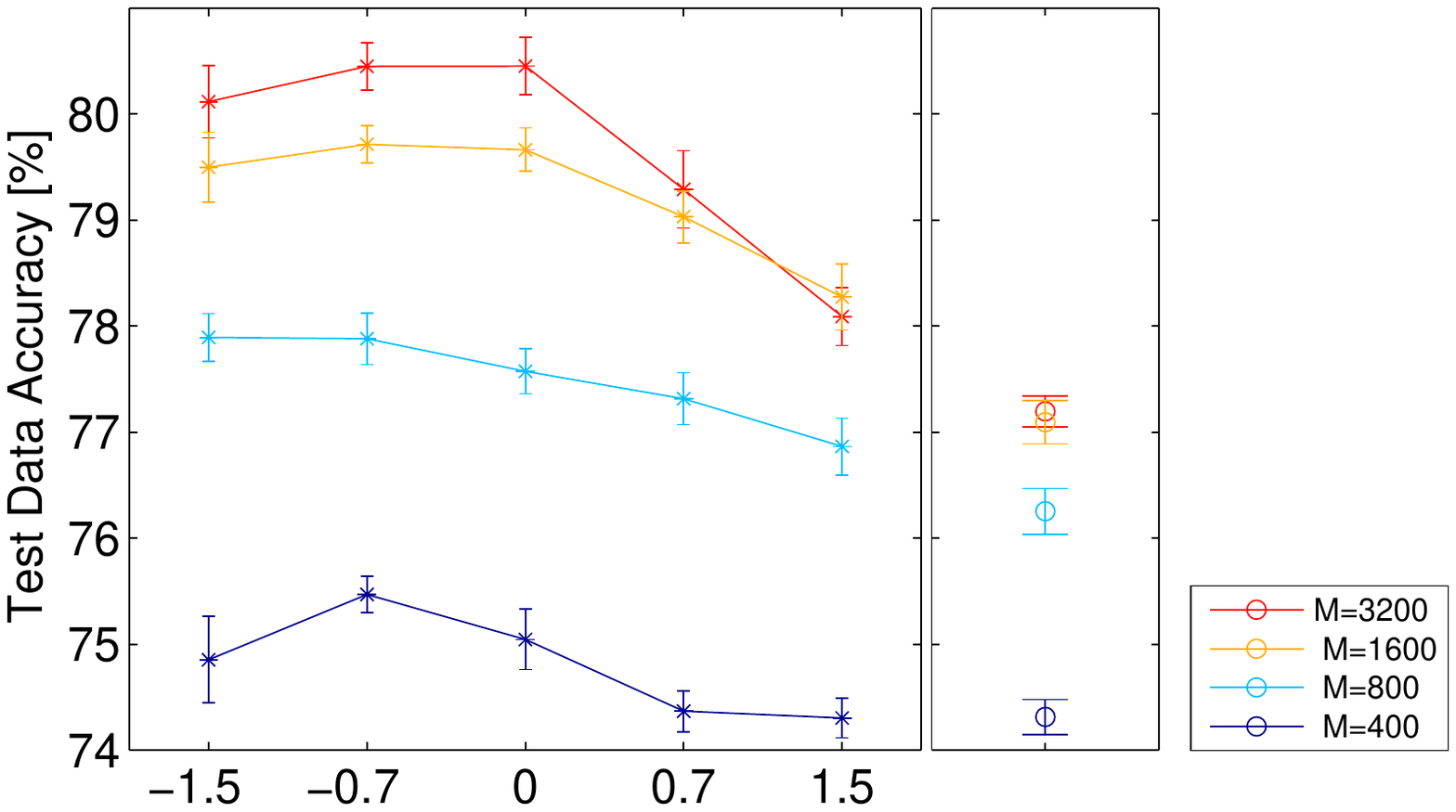}
\vspace{-43mm}
\caption*{\large\bf \hspace{-3mm}A\hspace{35mm}B}
\vspace{35mm}
\vspace{-5mm}
\caption*{\vspace{-4mm}\hspace{-7mm}$\theta$}
\vspace{2mm}
\caption{Performance and early stopping. Error bars indicate the standard deviation over 20 executions of training. Training settings from (c) to (e), and  \{3200, 108, 0.003, PCD\} were examined.
{\bf A}: A-infomax early stopping.
{\bf B}: Without early stopping. GRBMs trained for 80 epochs were used. }
\label{main:fig}
\end{minipage}
\end{tabular}
\end{figure}

 We can actually use AMI as a useful measure for GRBM representation quality. Plots (c) to (g) in \reffig{ADI_PFM:fig} for small receptive fields show that fine correlations between AMI and the test data accuracy are maintained regardless of the number of hidden units, the learning rate, and the weight update algorithm (Note that $\overline{\mathrm{AMI}}=-\mathrm{AMI}+\mathrm{Const}$. See appendix for more details). Plots (a) and (b) for larger receptive fields, however, do not show prominent correlations, reflecting that overcompleteness is not largely damaged. Even in these cases, drop in AMI is also observed. This phenomenon suggests that AMI is not a perfect measure for representation quality; AMI only serves as a good measure when the visible dimension is small as 100 in our experiments. Nevertheless, the high performance are often achieved with small receptive fields, i.e., small visible dimensions \cite{Coates:2011wo}. AMI, therefore, practically persists to be useful. 


\subsection{Performance Improvements by Early Stopping}

We next verify how a-infomax early stopping improves classification performance of stacked systems. We here only varied the number of hidden units and fixed the other training settings. We performed 20 runs of GRBM training for each of the settings. The threshold parameter for a-infomax early stopping (see appendix for more details) was selected as $\theta \in \{-1.5, -0.7, 0, 0.7, 1.5\}$. The GRBM training was terminated after 80 epochs, even if the early stopping condition did not hold. 

\refFig{main:fig} shows that a-infomax early stopping enables performance boosts in all the training settings shown. Moreover, there are two interesting trends to note. First, the larger the number of GRBM hidden units, the larger the performance gain by early stopping. Second, the larger the number of GRBM hidden units, the closer the timings at which performance and AMI reach their maximum values (this corresponds to performance peaks observed at $\theta\approx 0$ in \refFig{main:fig} A). 
 We can explain these trends that a larger gap between the total number of hidden units and the optimal number of effective filters in terms of likelihood maximization leads a severer and more obvious impact on performance that the lost of overcompleteness has. 
 These two trends are practically important, first because a larger number of hidden units generally results in higher classification accuracy \cite{Coates:2011wo}, and second because we can know that the optimal a-infomax early stopping thresholding parameter for GRBMs with a sufficiently large number of hidden units is $\theta\approx 0$, without tuning $\theta$. 
 The second property actually is the main advantage of a-infomax early stopping over our previous study where we needed to tune a thresholding parameter for GRBM reconstruction errors by cross validation \cite{Kiwaki:2013}. 

\begin{table}
\centering
\caption{CIFAR-10 test data classification accuracy by SVMs and unsupervised algorithms with the same number of hidden units. 20 GRBMs were trained under \{1600, 108, 0.003, CD\}. }
\label{pcomp:tab}
\begin{tabular}{|ll|c|}\hline
 Unsupervised Learning Algorithm & &Test Acc. \\\hline
GRBMs without early stopping & &$77.0\pm0.2$ \\
GRBMs with a-infomax early stopping  &($\theta=0$) &$\mathbf{79.7\pm0.2}$\\ \hline
 Sparse Coding &(\citet{Coates:2011uda})&78.9 \\ 
 OMP-1 &(\citet{Coates:2011uda})& 79.4\\ 
 OMP-10&(\citet{Coates:2011uda})& 80.1\\\hline
\end{tabular}
\end{table}

\subsubsection{Comparison with Other Approaches}
 A-infomax early stopping provides an approach for improving GRBM representations without extending the model. It has been proved that model extensions are effective for RBMs to gain overcomplete representations \cite{Courville:2011vh, Ranzato:2010vs}. However, these model extensions introduces some problems, such as conditional dependency of visible variables \cite{Ranzato:2010vs} or the need for a special treatment for the continuous hidden variables for stacking \cite{Courville:2011vh, Luo:2013vx}. A-infomax early stopping is free from such problems. Our approach thus will be directly applied to well-established deep learning models, such as deep Boltzmann machines \cite{Salakhutdinov:2007uk} or deep belief networks \cite{Hinton:2006tm}. Our findings may also help us to a deeper understanding of the mechanism by which such model extensions improve representation quality. 

The score achieved by GRBMs with a-infomax early stopping is comparable to the state-of-the-art score by single layer models that were reported to largely outperform GRBMs \cite{Coates:2011wo, Coates:2011uda} (in \reftab{pcomp:tab}). This striking result also demonstrates the impact of a-infomax early stopping and potential ability of GRBMs to model natural images. 


\section{Conclusions}
 We proposed a-infomax early stopping to enhance GRBM representations in terms of overcompleteness and data fitting. We reviewed a recently found phenomenon where a GRBM once gains and eventually loses sharp edge filters as the training proceeds. We attributed this phenomenon to a tradeoff between overcompleteness and fitting of GRBMs. Along this line, we developed a-infomax early stopping that enables GRBMs to gain representations that are overcomplete and fit data well. We performed experiments on stacks of a GRBM and an SVM to verify the classification performance improvement by a-infomax early stopping. We found huge performance boosts by a-infomax early stopping, and that performance can compete with the state-of-the-art performance by other single layer algorithms. 

\section*{Acknowledgments}
This research is supported by the Aihara Innovative Mathematical Modelling Project, the Japan Society for the Promotion of Science (JSPS) through the “Funding Program for World-Leading Innovative R\&D on Science and Technology (FIRST Program),” initiated by the Council for Science and Technology Policy (CSTP), and by JSPS Grant-in-Aid for JSPS Fellows (135500000216). We thank Mr. Sainbayar Sukhbaatar and Dr. Naoya Fujiwara for valuable discussion and helpful comments.


{\renewcommand{\baselinestretch}{-0.9}
\bibliographystyle{unsrtnat}
\bibliography{manu}}

\newpage
\appendix

\section{Techinical Details on A-infomax Early Stopping}
To detect the peak timing of AMI, we use following procedure. During RBM training, we check the loss of AMI relative to its peak value at time $T$: $\rli{T} =  {\mathrm{max}_{T'}\tee{T'}} - {\tee{T}}$. Because AMI become unimodal curve except for small fluctuations, simple thresholding of $\rli{T}$ results in two stopping time, one before and the other after the AMI peak timing. To avoid this difficulty, we use the following conversion $\rlii{T} =  \rli{T}\sign(T-{\argmin_{T'}\rli{T'}})$, where $\sign(\cdot)$ is a sign function; $\rlii{T}$ now shows a monotonically increasing shape. We perform a-infomax early stopping by thresholding $\rlii{T}$, 
\begin{align}
\argmin_{T}\left|\rlii{T} - \theta\right|,  
\end{align}
where $\theta$ is a threshold. As with REs, we first compute ${\mathrm{max}_{T'}\tee{T'}}$ and then determine the values of RBM parameters for supervised training. 

\section{Details of The Cross Validation}
 GRBMs were trained by PCD with the mini batch size being set to $100$, the learning rate being set as $\delta = 0.003$, and the number of hidden and visible units being set to 1,600 and 108, respectively.  
 We select the parameters as follows: the sparsity target $\ts$ of GRBMs was selected from 0.01 to 0.06 uniformly separated by 0.01, and the sparsity strength $\lambda$ was selected from 0.1 to 0.5 separated by 0.1. The SVM parameter $C$ was selected from $35, 75, 150, 300$ and the soft thresholding bias was selected from 0.1 to 0.7 separated by 0.1. 

\end{document}